\documentclass[10pt,twocolumn,letterpaper]{article}

\usepackage{iccv}
\usepackage{times}
\usepackage{epsfig}
\usepackage{graphicx}
\usepackage{amsmath}
\usepackage{amssymb}

\usepackage[breaklinks=true,bookmarks=false,colorlinks,linkcolor=red, citecolor=red]{hyperref}

\iccvfinalcopy 

\def\iccvPaperID{****} 

\ificcvfinal\fi

\begin{document}

\title{Semantic Segmentation on VSPW Dataset through Aggregation of Transformer Models}

\author{Zixuan Chen\\
Xiaomi Inc.\\
{\tt\small chenzixuan@xiaomi.com}
\and
Junhong Zou\\
Xiaomi Inc.\\
{\tt\small zoujunhong@xiaomi.com}
\and Xiaotao Wang\\Xiaomi Inc.\\{\tt\small wangxiaotao@xiaomi.com}}

\maketitle
\ificcvfinal\thispagestyle{empty}\fi

\begin{abstract}

Semantic segmentation is an important task in computer vision, from which some important usage scenarios are derived, such as autonomous driving, scene parsing, etc. Due to the emphasis on the task of video semantic segmentation, we participated in this competition. In this report, we briefly introduce the solutions of team ’BetterThing’ for the ICCV2021 - Video Scene Parsing in the Wild Challenge. Transformer is used as the backbone for extracting video frame features, and the final result is the aggregation of the output of two Transformer models, SWIN and VOLO. This solution achieves 57.35\% mIoU, which is ranked 3rd place in the Video Scene Parsing in the Wild Challenge. 

\end{abstract}

\section{The 1st Video Scene Parsing in the Wild Challenge}

Video Scene Parsing in the Wild(VSPW)\cite{miao2021vspw} is a newly released video semantic segmentation dataset with totally 3536 videos. Videos are with a frame rate 15 and is of an average length of about 5 seconds. Totally 124 kinds of things or stuffs are annotated. The aim of the challenge is to assign pre-defined semantic labels to pixels of all frames in videos of the test set of VSPW. The final evaluation metric is mIoU.

In this work, we perform a semantic segmentation task on VSPW dataset based on Transformers and model aggregation.

\section{Dataset Used}

Since Transformers usually need large amount of data for training, we include extra data for training in our work.Except VSPW training set, we use ADE20k, COCO and Cityscapes dataset in order to boost the performance of our model.

\subsection{VSPW dataset}

VSPW dataset is a video dataset with a frame rate of 15. It contains 3536 videos, with 251633 frames totally. Among these videos, 2806 videos with 198244 frames is set for training use, while other 343 videos for validation and 387 videos for test. Considering computational cost, all the images are resized to a resolution of 480p.

\subsection{COCO dataset}
\label{2.2}
We take the training set and validation set of the COCO-Stuff dataset for the use of training(about 123k images in all). COCO-Stuff is a large-scale image semantic segmentation dataset, with totally 172 classes. When using COCO dataset, we first remap the label of COCO onto the label of VSPW. Among all 172 classes, 131 classes have their corresponding class in VSPW, while for another 41 classes we just label then as 255(unlabeled). After remapping, we discard images that are not densely annotated (less than 80\% pixels annotated).This operation eliminates about 10k images.

\subsection{ADE20k dataset}
ADE20K is one of the most used semantic segmentation datasets which contains about 25K images that cover 150 different common categories . It includes 20K images for training, 2K images for validation, we use both as our training set. The data preprocessing is the same as the Sec \ref{2.2}.

\subsection{Cityscapes dataset}

CityScapes is one of the most used semantic segmentation datasets, which has 5000 high-quality annotated pictures of street scene in 19 classes. Since we noticed that there are fewer images of vehicles and poles in the original training set, we add it and performer preprocessing as Sec \ref{2.2}.

\section{Model Structure}

In general, we perform the semantic segmentation task through the aggregation of two models. The two models are both with an encoder-decoder structure, but with different backbone(Swin Transformer\cite{liu2021Swin} and VOLO\cite{yuan2021volo} separately). Meanwhile, the two models are trained under different conditions(training setting, dataset used, etc) to ensure the effective of model aggregation. 
\subsection{Transformer as Backbone network}

Since the appearance of fully convolutional network (FCN)\cite{long2015fully}, CNN has long dominated the field of semantic segmentation.
Most works on semantic segmentation use CNN, typically ResNet\cite{he2015deep} or HRNet\cite{WangSCJDZLMTWLX19}, as backbone network and design a proper decode head to utilize the output feature of the backbone, and finally get a dense predicting map.

However, in the field of natural language processing(NLP), Since the appearance of Transformer\cite{vaswani2017attention}, it has proved to be a milestone. Researchers have been trying to transplant the structure of Transformer into computer vision tasks. Vision Transformers(ViTs)\cite{dosovitskiy2021image} is the first work that uses Transformer in the task of image classification on ImageNet, achieving competitive result. After that, researchers have been trying to use Transformers in the task of semantic segmentation. For example, With ViTs served as backbone, SETR\cite{zheng2021rethinking} achieves a result of 48.64 mIoU on the ADE20k dataset, outperforming the models with CNN backbones.

We studied the state-of-the-art models on several datasets, especially Cityscapes and ADE20k. We find that Transformers performs much better on ADE20k, with SegFormer-B5\cite{xie2021segformer} achieving 51.8 mIoU, SETR achieving 48.64 mIoU. As a comparison, HRNet + OCR\cite{YuanCW19} achieves 47.98 mIoU on the validation set of ADE20k. 

While on Cityscapes, there appears an opposite result. HRNet + OCR performs the SOTA result with a hierarchical multi-scale aggregation, namely 85.1 mIoU on Cityscapes test set, higher than SegFormer-B5's 83.1 and SETR's 81.64 mIoU.

We consider that the difference is caused by the scale of dataset. ADE20k is a dataset with 150 classes and about 20000 images for training, while Cityscapes only has 19 classes and 3500 finely annotated images for training. Transformers may achieve better results on datasets with more classes and larger scale of data. Since VSPW is a large-scale video dataset with 124 classes and more than 190000 images for training, Transformers performs much better than CNN backbones. In our work, we choose Swin Transformer\cite{liu2021Swin} and VOLO\cite{yuan2021volo}, which separately achieves 53.5 and 54.3 mIoU on ADE20k, as the backbone network.

\subsection{Swin Transformer}

Vision Transformer performs self-attention among all the patches in each stage. While in Swin Transformer, the self-attention computation is limited in a set of non-overlapping windows. After compute self-attention in a local windows, all the windows will be shifted in the next self=attention computation so that connections between non-overlapping windows can be established.

Swin Transformer's design of shift window has 2 advantages. Since the amount of computation is fixed within a single window, Swin Transformer has linear complexity with respect to the size of image. By contrast, ViT has square complexity with respect to image size. In addition, the shift window enables Swin Transformer to construct hierarchical feature maps, which is suitable for dense prediction tasks like semantic segmentation.

\subsection{Vision Outlooker(VOLO)}
\cite{yuan2021volo} finds the major factor limiting VITs from outperforming CNNs is their low efficacy in encoding fine-level features into the token representations. To resolve this problem, \cite{yuan2021volo} introduce a novel outlook attention mechanism, termed Outlooker, to enrich the token representations and present a general architecture VOLO. The VOLO tokenizes images on patches and employs multiple Outlookers at fine level to improve the model performance. VOLO-D5 attains 84.3 mIoU score in Citysacpes validation set, and achieve 54.3 mIoU score in ADE20K validation set, largely improves the SOTA result. 

In this Challenge, we take VOLO-D5 as encoder, and through Unet architecture to restore resolution from 32x32 to 256x256, and finally take OCRNet as decoder following Unet. 

\subsection{OCRNet decoder}
OCRNet\cite{YuanCW19} has been proved to be a powerful decoder for semantic segmentation. With object attention module, OCRNet can explicitly enhance the representations of pixels with feature of the object that each pixel belongs to.

UperNet is often combined with backbones with a final output of stride 32 instead of OCRNet due to the low resolution. OCRNet is often combined with backbones that keep high resolution, such as HRNet and ResNet with dilated convolution. But when we experiment on VSPW dataset, we find that if we use a certain method to recover the resolution, OCRNet can achieve about the same performance as UperNet with less amount of computation. For example, for a model with Swin-L as backbone, when inputting a picture with size 480*853, using UperNet as decoder needs 640 GFlops, while OCRNet only needs 540 GFlops of computation.

\subsection{Structure details}
\label{3.5}
First, we train 2 models with encoder-decoder structure, separately using Swin Transformer and VOLO as backbone, and both using OCRNet as decoder.

The training of Swin Transformer is based on mmsegmentation\cite{mmseg2020}. For Swin Transformer, we choose Swin-Large as our backbone. The embed dim of the patch embedding module is 192. The depth of 4 stages are initially 2,2,18,2. 

The number of attention heads are 6,12,24,48. Swin Transformer extracts the feature of an image through 4 stages, thus have 4 outputs with stride 4,8,16,32. Output from lower stages is of higher resolution and low-level semantic information, while Output from higher stages have lower resolution and high-level semantic information. To make full use of the output of Swin Transformer, we take the last 3 outputs of Swin Transformer(stride 8,16,32), resize them to the resolution of stride 8, and perform a concatenating operation. After this step, we get a feature map with stride 8 and feature dim 2688. The stride 4 feature is discarded to prevent bringing to much amount of computation.

The feature map from Swin Transformer are both for FCN to generate coarse soft segmentation and for OCR to generate pixel representations. The pixel representation dim is set to 1024 and the number of middle channels in OCR's object context block is set to 512.

For Volo, we use VOLO-D5 given in the Volo's github repository as backbone. We use a U-net to combine the last 3 output feature maps and recover the resolution since we find such architecture is better than just resize-concatenate operation for Volo model. The OCRNet is set with almost the same parameters as Swin. 

\subsection{Model Aggregation}

Model aggregation indeed largely boosts the performance of our model. Since we train the 2 models in different conditions, they may give out different results on quite a lot of pixels. Combining the output result of 2 models can bring a progress of about +3 mIoU on both validation set and test set of VSPW. We perform model aggregation through a simple weighted summation.

\section{Implementation detail}

In this part we will describe the detailed training process of our models. 

\subsection{Training Settings}

\subsubsection{Swin Transformer Training Settings}
\label{swin_training_setting}
\ \ \ \ The initial model structure is Swin Transformer backbone + OCRNet decoder described in Sec \ref{3.5}, with ImageNet pretrained model downloaded from 

\href{https://github.com/microsoft/Swin-Transformer}{https://github.com/microsoft/Swin-Transformer}

The training process of Swin can be divided into 2 parts. Among the 2 parts, the way of data augmentation is shared, and batch-size is always set to 16. For data augmentation, we perform multi scale, random crop, random horizontal flip, metric distortion on images. 

Multi scale: 3 random variables(marked as $\alpha$,$\beta_1$,$\beta_2$) are used to determine the resize scale. $\alpha$ is first uniformly sampled from range (1.0,2.0), and with a 50\% probability to take its reciprocal.$\beta_1$ and $\beta_2$ are both uniformly sampled from range (-0.2,0.2). An image of size (h,w) is then resized to (h*$\alpha$*(1+$\beta_1$),w*$\alpha$*(1+$\beta_2$)).

Random Crop: since most of the picture from VSPW 480p dataset is of size (480,853), the crop size is directly set to (480,853).

Random horizontal flip: with a 50\% probability, a horizontal flip is applied on images.

Metric distortion: this augmentation includes 4 sub-operations, namely random brightness, random contrast, random saturation and random hue. Random brightness: pixels' RGB value are added with a random variable uniformly sampled from (-32,32). Random contrast: pixels' RGB value are multiplied with a random variable sampled from (0.5,1.5). Random saturation: convert an image to HSV color space, multiply the S value(saturation) with a random variable sampled from (0.5,1.5). Random hue: convert an image to HSV color space, add a value sampled from (-18,18) to the H value(hue) of the image. Each of the 4 operation is performed with a probability of 50\%.

Dataset used, training schedule and loss function are different among 2 parts. In the first part, we only use VSPW training set. An AdamW optimizer is used, with betas=(0.9,0.999).Weight decay of AdamW is set to 0.02. The model is trained for 160k iterations in this part. The learning rate follows a linear schedule and warm-up is applied. To be specific, the learning rate of backbone grow from 0 to 6e-6 in the first 1500 iterations, and gradually reduce to 0 in the rest iterations, learning rate is always changing in a linear way.The learning rate of decoder(OCRNet) is always 10 times of the backbone's learning rate. A pixel-distribution-based loss function(described in Sec \ref{4.2.1}) is used.

In the second part, we aim to boost the model's performance on some classes with low IoU, so we add COCO dataset for training. Still with AdamW optimizer, The model is trained for 40k iterations in this part. Without warm-up, both backbone and decoder take the learning rate that reduce from 1e-5 to 0 in a linear way. Further more, we deepen the model, with the depth of 4 stages added to (4,6,20,4). A confusion-matrix-based loss function(described in Sec \ref{4.2.2}) is used.

\subsubsection{VOLO Training Settings}
\ \ \ \ We get our baseline model VOLO-D5 architecture and pretrained models from \href{https://github.com/sail-sg/volo}{https://github.com/sail-sg/volo}. VOLO-D5 Model is trained with AdamW optimizer with default betas (0.95, 0.999) and uses imagenet pretrained model.The initial learning rate is set to 6e-5 and decreased by polynomial decay.


For data augmentation, we performer multi scale, random flip, and color jitter on images. Multi scale setting is the same as \ref{swin_training_setting}. Besides, after the model converges, we will train for a few additional epochs, and average the weights of these models. Finally a simply cross entropy loss with a class weight in pytorch(described in \ref{4.2.1}) is used in training process.

\subsection{Loss}

\subsubsection{Pixel-distribution-based Loss}
\label{4.2.1}
\ \ \ \ The problem of class unbalance appears in VSPW dataset, so we calculate a class weight for each class. We count the number of pixels belonging to 124 classes in training set of VSPW, marked from $n_1$ to $n_{124}$. The weight of class i is $\sqrt{n_i/\mu_n}$, where $\mu_n$ is the average of $n_i$.

We design a loss function based on the cross entropy loss(CE) and class weight. In Pytorch, CE with class weight is computed through 

$Loss = \frac{1}{n}\sum_{i=0}^n w_{y_i} \ln{p_{y_i}} $

where $y_i$ is the ground truth of the $i^{th}$ pixel, $p_{y_i}$ is the probability predicted by the segmentor, $w_{y_i}$ is the class weight of class $y_i$. But we find that if a pixel of classes with low weight is wrongly recognized as classes with high weight, it will only produce a very small loss —— but this will largely do harm to the IoU of classes with small weight. So we change the loss function to

$Loss = \frac{1}{n}\sum_{i=0}^n -max(w_{y_i},w_{y_i'}) \ln{p_{y_i}} $

where $y_i$,$y_i'$ is respectively the ground truth and predicted class of the $i^{th}$ pixel.

\subsubsection{Confusion-matrix-based Loss}
\label{4.2.2}
\ \ \ \ This loss function is based on focal loss,letting model focus on pixels with low confidence, and is used to finetune Swin Transformer for a higher mIoU. We first compute a confusion matrix $\{C_{ij}\}$ on validation set, where $C_{ij}$ represents the number of pixels that have ground truth class i and is predicted as class j. The loss function is defined as

$Loss = \frac{1}{n}\sum_{i=0}^n -\frac{C_{y_i y_i'}}{min(C_{y_i y_i},C_{y_i' y_i'})} (1-p_{y_i})^2 \ln{p_{y_i}} $

\subsection{Test Time Augmentation}

Test time augmentation is an effective method to improve the performance of models. We add multi-scale input and flipping during test time. For Swin Transformer, we find that the model can finely adapt to input images of different size. So we resize test images with scale 0.5,1.0,1.5, and do flipping operation as test time augmentation. For VOLO, we find that the model is very sensitive to multi-scale input. Inputting images with a scale different from training will do harm to the performance. So we just resize test images to the training crop size and do flipping as test time augmentation. 

\subsection{Model Aggregation Detail}
\label{4.4}
After test time augmentation, the Swin Transformer model and the Volo model will both output a soft classification result(a matrix with shape $124*h*w$. We find that simply applying a weighted summation on the soft classification result can greatly improve the model performance. Let an aggregation param be $\gamma$, then the final output soft classification result is

$P = \gamma P_s + (1-\gamma)P_v$

where $P_s$ is the soft classification result of Swin Transformer and $P_v$ is Volo's.

We find that Swin Transformer and Volo achieve similar performance on the test set(both 51.59 mIoU when we first tried model aggregation), so set $\gamma$ to 0.5 is often a easy but effective choice. In the end of the competition, we finetune $\gamma$ to 0.56 according to the result on test set.

\section{Experiment Result}

In this part we will show the performance of our model on VSPW validation and test set. 

\subsection{Swin Model Result}

Several factors affect the final model performance. In our experiment, most influential factors include the model structure, crop-size, extra dataset and test time augmentation.Detailed experiment result in shown in Tab \ref{table 1}

All with OCRNet as decoder, models with CNN based backbones, ResNet-101 and HRNet-W48 only achieve lower than 40 mIoU on validation and test set. The Swin-L backbone brings about an improvement of more than 12 mIoU. This may partly attribute to the large size of model, but we can still know about the power of Transformer on a large-scale dataset.

The change of crop size brings about +2 mIoU, since most images from VSPW(480p) is of the size (480,853). Let crop size be the same as image size ensures the models to perform best during inference.

The adding of COCO dataset enhances models' performance against classes that cannot be well recognized. For example, on the class 'food', before adding COCO, our models can only achieve an mIoU of less than 10 on validation set. After adding COCO, the performance improves to 25.09.

Test time augmentation also brings benefit of +1.1 mIoU in all.

\begin{table}
\centering
\begin{tabular}{ccc}

\hline  
models & val mIoU & test mIoU\\
\hline  
ResNet101+OCRNet&36.68&34.02\\
\hline 
HRNet-W48+OCRNet& - &37.26\\
\hline 
Swin-L+OCRNet\\+crop size 512*512&53.79&49.47\\
\hline 
Swin-L+OCRNet\\+crop size 480*853&55.70&51.59\\
\hline 
Swin-L+OCRNet\\+crop size 480*853\\+COCO-dataset&56.80& - \\
\hline 
Swin-L+OCRNet\\+crop size 480*853\\+COCO-dataset\\+flip&57.26& - \\
\hline 
Swin-L+OCRNet\\+crop size 480*853\\+COCO-dataset\\+flip\\+multi scale
&57.89& - \\
\hline 

\end{tabular}
\caption{the model performance on validation and test set with different training settings and data augmentation}
\label{table 1}

\end{table}

\subsection{Volo Model Result}
The performance of Volo is shown in Tab \ref{table 2}. In VOLO Experiment, we get the baseline model with 48.65 mIoU, then the multi-scale resize was added to the training process, then got a 1.24 mIoU improvement. After the training process converges, we pick out 4 model weight saved during training and use the average of their weights as a new model. The new model achieves 50.77 mIoU. After that, changing input resolution to 480*720 and adding U-net to the net structure provide +0.52 and +0.30 mIoU respectively. The model with 51.59 mIoU is the first Volo model to be used for model aggregation, so most of the rest experiment is on aggregated models.
\begin{table}

    \centering
    \begin{tabular}{ccc}
    \hline
    experiment & resolution &test mIoU  \\
    \hline
     volo + ocr &  512x512 & 48.65  \\ 
     \hline
     volo  + ocr \\ + multiscale &  512x512   &  49.89\\
     \hline
     volo  + ocr \\ + multiscale \\+ weight average&  512x512  & 50.77 \\
    \hline
    volo  + ocr \\ + multiscale \\+ weight average &  720x480 & 51.29 \\
    \hline
    volo + unet + ocr \\ + multiscale \\+ weight average&  720x480  & 51.59\\
    \hline
    \end{tabular}
    \caption{VOLO model experiment }
    \label{table 2}
\end{table}

\subsection{Model Aggregation Result}
We perform model aggregation after both Swin and Volo have acceptable results on test set. The first attempt of model aggregation is performed when both models reach an mIoU of 51.59 on test set(line 4 in Tab \ref{table 1} and line 5 in Tab \ref{table 2}). Here we simply choose 0.5 as the aggregation parameter $\gamma$ (described in Sec \ref{4.4}). The mIoU directly reaches 55.10. Model aggregation can be performed together with test time augmentation and get a stronger result. Test time augmentation brings the result to 56.78 mIoU. Extra datasets are included to finetune both models. COCO is added to the training of Swin Transformer, while ADE20k and Cityscapes are added to train VOLO, and the extra datasets make mIoU improve to 57.83. Finally, by finely, searching $\gamma$, we choose a $\gamma$ of 0.56 and finally get the result 58.06 mIoU on test set. Here 58.06 is computed in the development phase, with a half of test set. The performance is 57.35 mIoU on the whole test set, which is our score in final phase. The experiment result is briefly shown in Tab \ref{table 3}.

\begin{table}
\centering
\begin{tabular}{cc}

\hline  
models &  test mIoU\\
\hline  
Swin-L(Line 4 in Tab \ref{table 1}) \\+ Volo(line 5 in Tab \ref{table 2}) &55.10\\
\hline 
Swin-L + Volo \\+ test time augmentation &56.78\\
\hline 
Swin-L(COCO finetune) \\+ Volo(ADE20k+Cityscapes finetune) \\+ test time augmentation &57.83\\
\hline 
Swin-L(COCO finetune) \\+ Volo(ADE20k+Cityscapes finetune) \\+ test time augmentation \\+ $\gamma$ set to 0.56 &58.06\\
\hline 

\end{tabular}
\caption{the result of model aggregation}
\label{table 3}

\end{table}

\section{Conclusion}
Transformers have shown their power as feature extractor in a semantic segmentation task. In our implementation, we mainly boost our performance by finding ways to train better Transformer-based models.

What we feel regretful is that we failed to find a proper way to leverage the temporal correlation between video frames. We have tried TMANet\cite{wang2021temporal} to build temporal-attention among frames. In addition, we have tried Video Swin Transformer\cite{liu2021video} to directly extract the feature of a frame sequence. However, neither of these methods shows advantage over methods that directly process single frame.

{\small
\bibliographystyle{ieee_fullname}
\bibliography{egbib}
}

\end{document}